\documentclass[letterpaper]{article} %
\usepackage[]{aaai25}  %
\usepackage{times}  %
\usepackage{helvet}  %
\usepackage{courier}  %
\usepackage{xcolor}
\usepackage[hyphens]{url}  %
\usepackage{graphicx} %
\usepackage{natbib}  %
\usepackage{caption} %
\usepackage{algorithm}
\usepackage{algorithm}
\usepackage{algorithmic}

\newcommand{\myheading}[1]{\vspace{.75ex}\noindent \textbf{#1}}
\usepackage{marvosym}
\usepackage{amssymb}
\usepackage{amsfonts}
\usepackage{amsmath}
\usepackage{booktabs}
\usepackage{multirow}
\usepackage{xcolor, colortbl}
\usepackage[capitalize]{cleveref}
\crefname{section}{Sec.}{Secs.}
\Crefname{section}{Section}{Sections}
\Crefname{table}{Table}{Tables}
\crefname{table}{Tab.}{Tabs.}
\usepackage{tikz}
 
\definecolor{mydarkblue}{rgb}{0,0.08,1}
\definecolor{mydarkgreen}{rgb}{0.02,0.6,0.02}
\definecolor{myred}{rgb}{1.0,0.0,0.0}
\definecolor{myred2}{rgb}{0.7,0.1,0.1}
\definecolor{mydarkblue2}{rgb}{0.05,0.1,0.7}
\definecolor{mypurple}{rgb}{111,0,255}
\definecolor{mypurple2}{rgb}{111,0,111}

\usepackage{pifont}%

\def\Approach{SwiftTry}

\usepackage{newfloat}
\usepackage{listings}
\DeclareCaptionStyle{ruled}{labelfont=normalfont,labelsep=colon,strut=off} %
\floatstyle{ruled}
\newfloat{listing}{tb}{lst}{}
\floatname{listing}{Listing}
\def\dataname{TikTokDress}

\title{SwiftTry: Fast and Consistent Video Virtual Try-On with Diffusion Models}
\author {
    Hung Nguyen\equalcontrib,
    Quang Qui-Vinh Nguyen\equalcontrib,
    Khoi Nguyen,
    Rang Nguyen
}
\affiliations {
    VinAI Research, Vietnam \\
    \{v.hungnm66, v.quangnqv, v.khoindm, v.rangnhm\}@vinai.io
}

\usepackage{bibentry}

\begin{document}

\maketitle

\begin{abstract}
    Given an input video of a person and a new garment, the objective of this paper is to synthesize a new video where the person is wearing the specified garment while maintaining spatiotemporal consistency. Although significant advances have been made in image-based virtual try-on, extending these successes to video often leads to frame-to-frame inconsistencies. Some approaches have attempted to address this by increasing the overlap of frames across multiple video chunks, but this comes at a steep computational cost due to the repeated processing of the same frames, especially for long video sequences. To tackle these challenges, we reconceptualize video virtual try-on as a conditional video inpainting task, with garments serving as input conditions. Specifically, our approach enhances image diffusion models by incorporating temporal attention layers to improve temporal coherence. To reduce computational overhead, we propose ShiftCaching, a novel technique that maintains temporal consistency while minimizing redundant computations. Furthermore, we introduce the \dataname~dataset, a new video try-on dataset featuring more complex backgrounds, challenging movements, and higher resolution compared to existing public datasets. Extensive experiments demonstrate that our approach outperforms current baselines, particularly in terms of video consistency and inference speed. The project page is available at \url{https://swift-try.github.io/}.
\end{abstract}

\section{Introduction}
\begin{figure}
  \centering
  \includegraphics[width=1\linewidth]{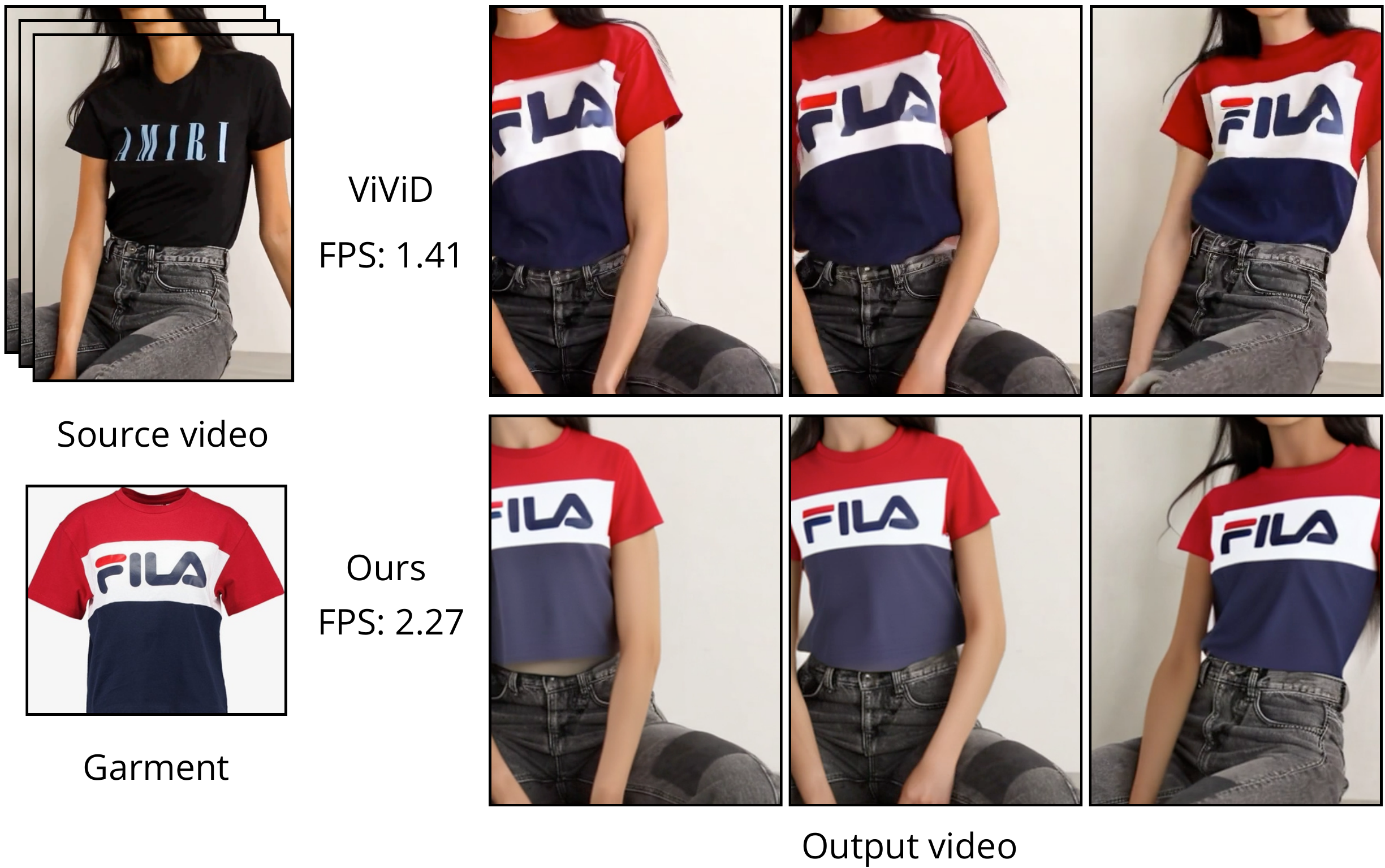}
      \caption{Results of our \Approach~compared to those of ViViD \cite{fang2024vivid}, a previous method for video try-on. Our method preserves garment texture detail and consistency while achieving over 60\% faster runtime.}

   \label{fig:teaser}
\end{figure}

Video virtual try-on is an emerging research area~\cite{chen2021fashionmirror, rogge2014garment, pumarola2019unsupervised, dong2019fw, kuppa2021shineon, zhong2021mv, jiang2022clothformer, he2024wildvidfit, xu2024tunnel, fang2024vivid, zheng2024viton} with significant potential in fashion and e-commerce. The ability to realistically visualize how a garment appears on a person in a video could transform online shopping. However, despite recent advances in image-based virtual try-on~\cite{he2022style, choi2021viton, lee2022hrviton, xie2023gp, zhu2023tryondiffusion, kim2023stableviton}, extending these capabilities to video remains challenging due to the need for spatiotemporal consistency and the high computational costs of processing long sequences.

A significant challenge in video virtual try-on is balancing the need for temporal coherence with the computational demands of processing long video sequences. Previous methods~\cite{xu2024tunnel, he2024wildvidfit, fang2024vivid} often struggle with temporal inconsistencies, resulting in visual artifacts and flickering between frames, which undermines the realism of the virtual try-on experience. Additionally, the high computational cost of rendering high-quality results over extended sequences limits the practicality of these approaches in real-world applications.

Another challenge is the lack of an adequate evaluation dataset. The first public video try-on dataset, VVT~\cite{dong2019fw}, only includes basic pattern garments, form-fitting T-shirts, uniform backgrounds, static camera angles, and repetitive human motions. More recently, ViViD~\cite{fang2024vivid} introduced the first practical dataset for video virtual try-on. However, it struggles to handle in-the-wild scenarios, such as complex movements and diverse backgrounds, making it difficult to meet the demands of real-world applications. Moreover, the poor quality of video try-on results can often be attributed to the inaccurate masks extracted using human parsing segmentation \cite{li2020self}, which are applied to each frame of the video.

In this paper, we address these challenges with two key contributions. First, we introduce a new high-quality dataset, named \textbf{\dataname}, consisting of 817 videos specifically designed for training and evaluating video virtual try-on models. This dataset features realistic scenes, diverse garment types, and complex movements, providing a robust foundation for advancing research in this field. Second, we propose a novel video virtual try-on framework named \textbf{\Approach}, as illustrated in \cref{fig:teaser}, which significantly reduces the computational cost of processing long video sequences while maintaining temporal consistency.

Our framework is inspired by state-of-the-art diffusion-based image virtual try-on methods~\cite{kim2023stableviton, xu2024ootdiffusion, choi2024improving} and incorporates temporal attention within the UNet architecture to train on video try-on data. During inference, we introduce a new technique called \textbf{ShiftCaching}, which ensures temporal coherence and smooth transitions between video chunks while minimizing redundant computation compared to previous methods. Extensive experimental results demonstrate that our proposed \Approach~framework, leveraging these techniques, significantly outperforms existing video virtual try-on methods in both accuracy and efficiency.

In summary, the contributions of our work are as follows:
\begin{itemize}
    \item We propose a new  technique for video inference named ShiftCaching, which can ensure temporal smoothness between video chunks and reduce redundant computation.
    \item We introduce and curate a new video virtual try-on dataset, \dataname, which encompasses a wide range of backgrounds and complex movements and features high-resolution videos, filling a gap that exists in previous video virtual try-on datasets.
\end{itemize}

\begin{figure*}[t]
  \centering
  \includegraphics[width=1\linewidth]{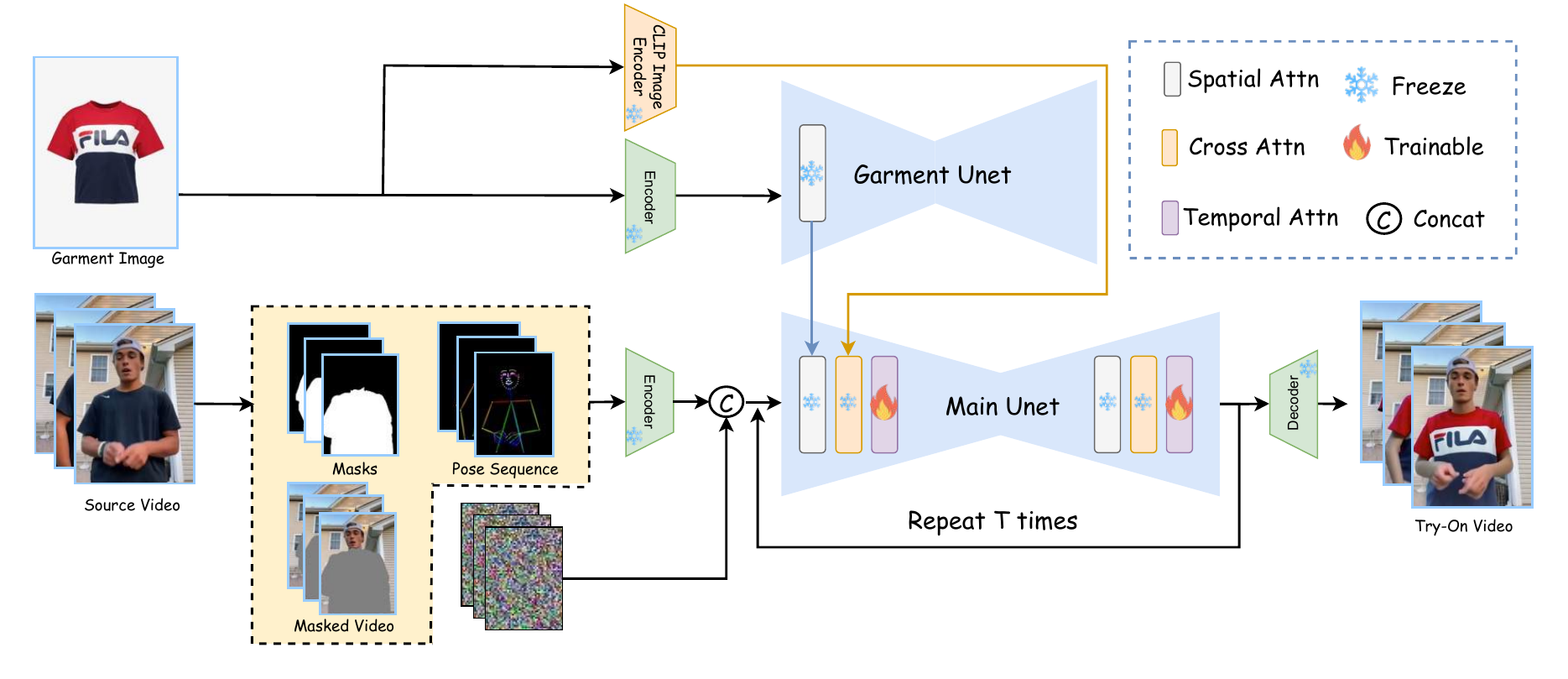}
       \caption{Overview of Stage 2 of our \Approach~framework (Note that stage 1 is similar, except the input is a single image frame, and it does not include temporal attention layers). Given an input video and a garment image, our method first extracts the masked video, corresponding masks, and pose sequence. The masked video is encoded into the latent space by the VAE Encoder, which is then concatenated with noise, masks, and pose features before being processed by the Main U-Net. To inpaint the garment during the denoising process, we use a Garment U-Net and a CLIP encoder to extract both low- and high-level garment features. These features are integrated into the Main U-Net through spatial and cross-attention mechanisms.}
   \label{fig:overview}

\end{figure*}

\section{Related Work}
\myheading{Image Virtual Try-On.}
Traditional image virtual try-on methods \cite{han2018viton, wang2018toward, Dong_2019_ICCV, Yang_2020_CVPR, ge2021parser, he2022style, choi2021viton, lee2022hrviton, xie2023gp} commonly employ a two-stage pipeline based on \textbf{GANs} \cite{goodfellow2014generative}.  In this approach, the target clothing is first warped and then fused with the person image to create the try-on effect. Various techniques have been utilized for clothing warping, including thin-plate spline (TPS) warping \cite{han2018viton}, spatial transformer networks (STN) \cite{li2021toward}, and flow estimation \cite{xie2023gp}. Despite these advances, such methods often face limitations in generalization, resulting in significant performance degradation when applied to person images with complex backgrounds.

Recently, \textbf{diffusion models} have markedly enhanced the realism of images in generative tasks, leading to their increasing adoption in virtual try-on research. For instance, TryOnDiffusion \cite{zhu2023tryondiffusion} presents a virtual try-on method utilizing two U-Nets, but it requires a large dataset of image pairs of the same person in different poses, which can be difficult to acquire.  StableVITON \cite{kim2023stableviton} conditions the garment in a ControlNet \cite{zhang2023adding}-style using a zero cross-attention block, while IDM-VTON \cite{choi2024improving} proposes GarmentNet to encode low-level features combined with high-level semantic features extracted via IP-Adapter \cite{ye2023ip}. Despite these advancements, extending these existing image virtual try-on methods for video often results in significant inter-frame inconsistency and flickering, which adversely affects the overall quality of the generated results.

\myheading{Video Virtual Try-On.}
Several efforts have been made to develop virtual try-on systems for videos. FW-GAN~\cite{dong2019fw} incorporates an optical flow prediction module from Video2Video~\cite{wang2018video} to warp preceding frames to the current frame, enabling the synthesis of temporally coherent subsequent frames. MV-TON~\cite{zhong2021mv} introduces a memory refinement module that retains and refines features from previous frames. ClothFormer~\cite{jiang2022clothformer} employs a vision transformer in its try-on generator to minimize blurriness and temporal artifacts. It also features an innovative warping module that combines TPS-based and appearance-based methods to address challenges such as incorrect warping caused by occlusions. Among \textbf{diffusion-based} methods, Tunnel Try-On~\cite{xu2024tunnel} is the first to apply diffusion models for video virtual try-on, effectively handling camera movement and maintaining consistency. However, its demo videos are limited to only a few seconds in length. ViViD~\cite{fang2024vivid} introduced a large-scale video try-on dataset with multiple categories, but it remains limited by simple backgrounds and movements, which constrain its ability to ensure long-term consistency and coherence. In this paper, we propose a novel approach that establishes temporal smoothness and coherence across video chunks. Additionally, we integrate a caching technique~\cite{ma2024deepcache} to reduce redundant computations during long video inference, significantly improving efficiency.

\section{Methods}
\label{sec:method}

\myheading{Problem Statement:}
Given a source video $V = \{I_1, I_2, \dots, I_N\} \in \mathbb{R}^{N \times 3 \times H \times W}$ of a person and a garment image $ g \in \mathbb{R}^{3 \times H \times W} $, where \( N \), \( H \), and \( W \) represent the video length, frame height, and frame width, respectively, our goal is to synthesize a target video $ \hat{V} = \{\hat{I}_1, \hat{I}_2, \dots, \hat{I}_N\} \in \mathbb{R}^{N \times 3 \times H \times W}$  of the person wearing the garment, while preserving the motion of the person, the background in $ V $, and the color and texture of $ g $.

It is important to note that collecting both source and target videos of the same person with identical motions and gestures, differing only in the garment, is extremely challenging. As a result, most video try-on approaches adopt a \textbf{self-supervised training} method, where only a single video is used, and the garment regions are masked. The model is then trained to inpaint the masked regions using guidance from the garment image.

In the next section, we first describe our overall video try-on architecture and then discuss in detail the ShiftCaching technique -- one of our main contributions.

\begin{figure}[t]
    \centering
    \includegraphics[width=1\linewidth]{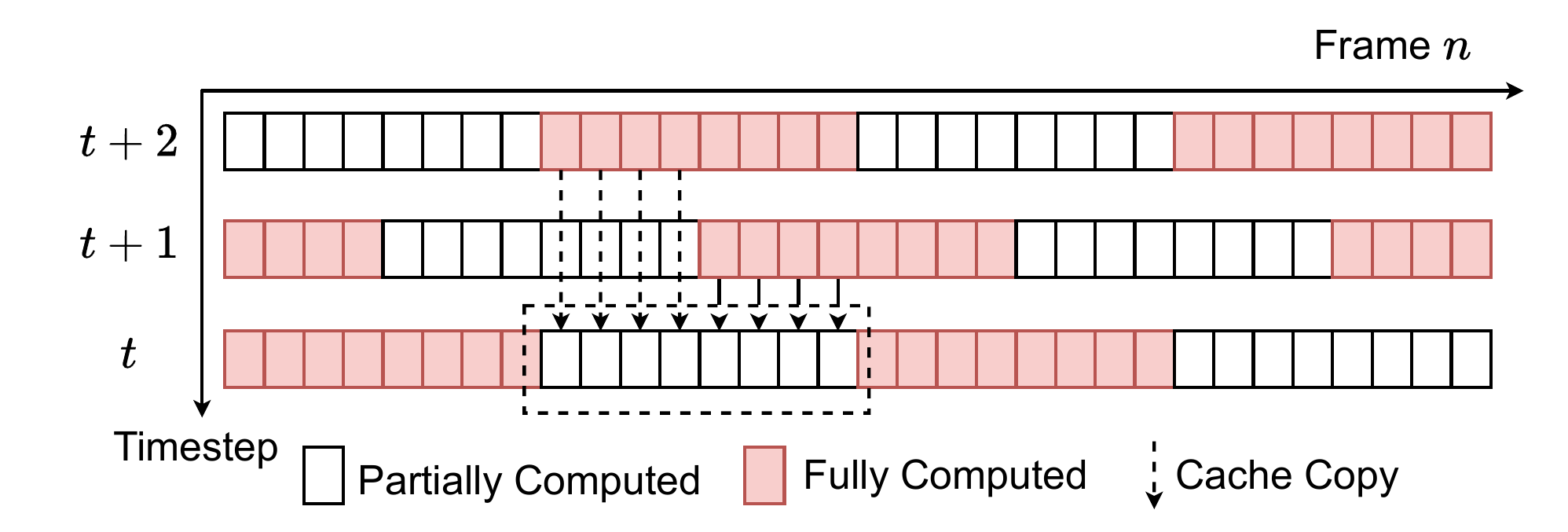}
    \caption{Illustration of fully and partially computed frames with a chunk size of $N = 8$ and a shift of $\Delta = 4$. In the partially computed chunk, one half uses cached features from $t+2$, while the other half uses features from $t+1$.}
    \label{fig:shiftcaching_overview}
\end{figure}

\subsection{Overall Architecture}
Our approach comprises two stages: first, training a diffusion-based image try-on model, and second, extending it to work with video data by incorporating temporal attention into every block of the Main UNet.

In the first stage, inspired by StableVITON~\cite{kim2023stableviton}, we design a \textbf{diffusion-based image try-on model} with two submodules: the Garment UNet and the Main UNet, as illustrated in \cref{fig:overview}. The \textit{Main UNet} is a modified inpainting model initialized with pretrained weights from Stable Diffusion~\cite{rombach2022high}. It takes as input four channels of latent noise, four channels of latent representations of the masked image (i.e., the person image with the clothing region masked), and one channel for the binary mask representing the inpainting region. To further enhance generation quality, we add the pose skeleton as an additional control, represented by a pose map rendered from DW-Pose~\cite{yang2023effective}. This results in a 13-channel input, which is fed into the Main UNet to predict the cleaned latent over $T$ timesteps. Finally, the cleaned latent is passed through a decoder to produce the output image.

The \textit{Garment UNet} has a similar architecture to the Main UNet but only takes the garment image as its input, rather than the multiple channels used in the Main UNet. This module is designed to extract both detailed and high-level features from the garment, guiding the Main UNet to accurately replicate the garment’s appearance through Reference Attention. Specifically, we follow the Reference Attention mechanism from AnimateAnyone~\cite{hu2023animateanyone}, replicating the Garment UNet’s feature maps along the temporal dimension and concatenating them with the Main UNet’s feature maps along the spatial dimension before applying the UNet’s self-attention. We use the VITON-HD dataset~\cite{choi2021viton} to train our network during this stage.

\begin{table}[t]
\centering
\small
\setlength{\tabcolsep}{4pt}

\begin{tabular}{ccc}
  \toprule
    \textbf{Overlapping size} &   \textbf{VFID$_{\text{I3D}}$ $\downarrow$} & \textbf{FPS $\uparrow$} \\
  \midrule
  $S=0$ & 9.040 & \textbf{1.544}  \\
  $S=4$  &  8.822 & 1.176 \\
  $S=8$ &  8.947 &  0.801 \\
  $S=15$ & \textbf{8.675} & 0.104 \\
  \bottomrule
\end{tabular}
\caption{Trade-off between speed (FPS) and consistency (VFID$_{\text{I3D}}$) with different overlap sizes of previous methods.}
\label{tab:overlap}
\end{table}

\begin{figure}[t]
    \centering
    \includegraphics[width=1\linewidth]{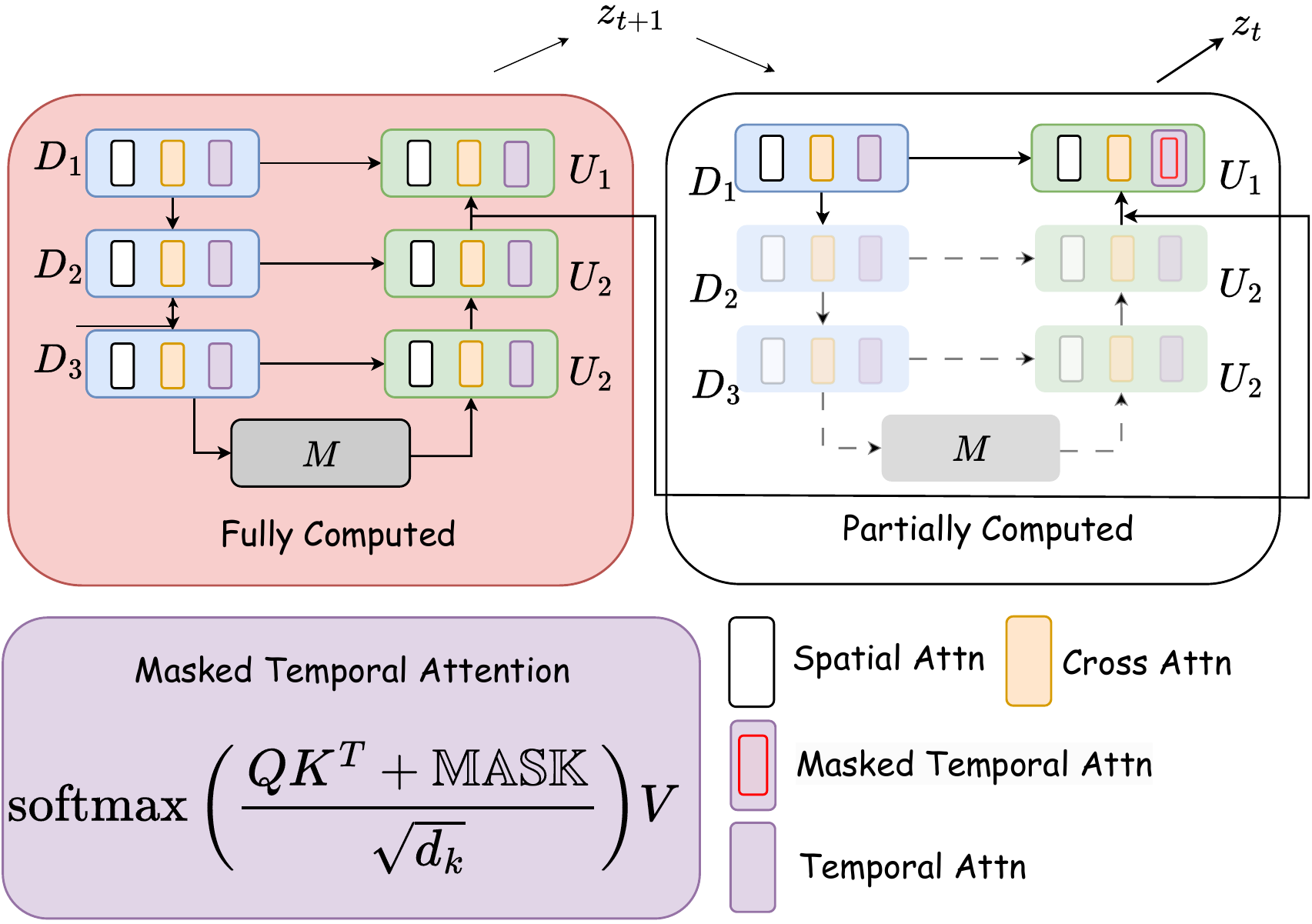}
    \caption{Comparison between a fully computed frame and a partially computed frame. The partially computed frame employs Masked Temporal Attention instead of standard Temporal Attention to resolve mismatches in cached features.}
    \label{fig:ShiftCaching2}
\end{figure}

In the second stage, we modify the Main UNet from the image try-on model to ensure temporal consistency across video frames. This adaptation involves converting its 2D layers into pseudo-3D layers \cite{guo2023animatediff, wu2023tune, zhou2022magicvideo} and adding a temporal attention layer after the Spatial and Cross-Attention layers to capture temporal correlations between frames. The architecture of the modified UNet blocks is illustrated in \cref{fig:overview}. In the temporal attention layer, the features are reshaped into the shape of  $(H \times W) \times N \times C$, where  $C$  is the number of feature channels and $H\times W$ is the batch-size dimension, to compute self-attention along the temporal dimension. This allows a location in the latent space of frame $t$  to interact with the same location in other frames within a chunk of $N$ frames. This design is highly efficient as it avoids the expense of full 3D attention by factorizing it into two consecutive steps: spatial attention (2D) and temporal attention (1D). This approach allows a location in one frame to exchange information with every location in all other frames.
Additionally, we incorporate sinusoidal positional encoding to help the model recognize the position of each frame in the video, following \cite{guo2023animatediff}. In this stage, we train only the temporal attention layer while keeping the other layers unchanged, using a video dataset.

\subsection{ShiftCaching Technique}
\label{shift_caching}

Due to memory constraints, current video diffusion-based virtual try-on methods can only generate video chunks of 16 frames at a time. Previous approaches \cite{fang2024vivid, he2024wildvidfit, xu2024tunnel} use a temporal aggregation technique \cite{tseng2023edge, xu2023magicanimate} to stitch overlapping video chunks into longer sequences. In this process, the long video is divided into overlapping chunks with an overlap size $S$, typically set to $N/2$ or $N/4$. At each denoising timestep $t$, the overlapping noise predictions are merged using a simple averaging technique. However, this method involves a trade-off: a smaller overlap size, such as $S=4$, can cause temporal flickering and texture artifacts, while a larger overlap size, such as $S=15$, improves consistency but greatly slows down the process as shown in \cref{tab:overlap}.

To achieve good temporal coherence and smoothness without recomputing the overlapped regions, we propose a \textbf{shifting mechanism} during inference. Specifically, we divide the long video into non-overlapping chunks ($S=0$). At each DDIM sampling timestep $t$, we shift these chunks by a predefined value $\Delta$ between two consecutive frames, allowing the model to process different compositions of noisy chunks at each step. An example of a fixed $\Delta=4$ applied to a chunk with length $N=8$ is illustrated in \cref{fig:shiftcaching_overview}.

To further accelerate the inference process, we can skip a random chunk to reduce redundant computation during denoising. However, naively dropping chunks without adjustment can lead to abrupt changes in noise levels in the final results. Following~\cite{ma2024deepcache}, which notes that adjacent denoising steps share significant similarities in high-level features, we instead perform \textbf{partial computations} on the Main U-Net. Specifically, we use a cache to copy the latest features from the fully computed timestep $z_{t+1}$ (Red frame) and use these features to partially compute the current latent $z_t$ (White frame), bypassing the deeper blocks of the UNet, as illustrated in \cref{fig:ShiftCaching2}.

When performing partial computations on a chunk, the cached features typically include the first half from timestep $t+2$ and the second half from timestep $t+1$, which can lead to mismatches between the two halves. To address this, we introduce a \textbf{Masked Temporal Attention} mechanism. This mechanism applies a special mask of size $N \times N$ during the softmax attention calculation to set specific values in the attention matrix to $0$. This prevents the transfer of information from less accurate features (timestep  $t+2$) to more accurate features (timestep  $t+1$), while allowing transfer from good features to bad features. This approach ensures both smoothness and high quality in the partially computed cells.

\section{\dataname~Dataset}

\begin{table*}[t]
\centering
\small
\setlength{\tabcolsep}{5pt}
\begin{tabular}{c c c c c c c c c c c c c c c c}
  \toprule
  \multirow{2}{*}{\textbf{Label}} & \multicolumn{2}{c}{\textbf{Gender}} & \multicolumn{3}{c}{\textbf{Skin tone}} & \multicolumn{3}{c}{\textbf{Camera position}} & \multicolumn{2}{c}{\textbf{Distance}} & \multicolumn{2}{c}{\textbf{Action}} & \multicolumn{2}{c}{\textbf{Background}}  \\ 
  \cmidrule{2-15}
  & Male & Female & White & Asian & Black & Bottom & Top & Center & Near & Far & Move & Stay & Dynamic & Static\\
  \midrule
  Counting & 267 & 550 & 541 &124 & 152 & 576 & 7 & 231 & 570 & 247 & 275 & 542 & 94 & 724\\ 
  \bottomrule
\end{tabular}
\caption{Data statistics highlighting the diversity and complexity of our dataset. The table provides a breakdown of attributes such as gender, skin tone, camera positions, distances (Near/Far), and actions (Move/Stay), indicating whether the actor is moving or stationary}

\label{tab:datastat}
\end{table*}

\begin{figure}[t]
\centering
\includegraphics[width=1\linewidth]{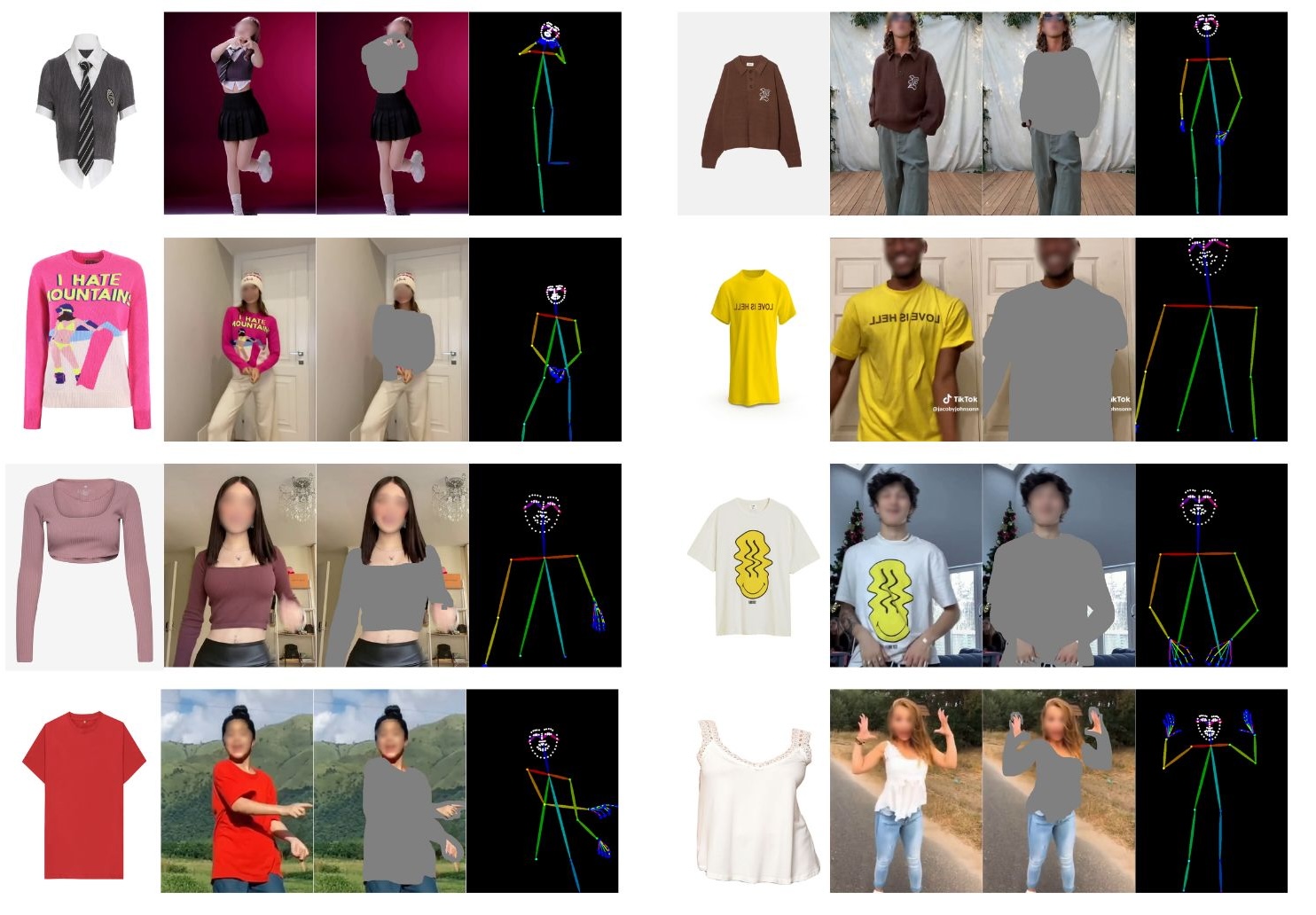}
\caption{Example videos from the \dataname~dataset highlighting diversity in skin tones, genders, camera angles, and clothing types.}
\label{fig:dataset_samples}
\end{figure}

\begin{figure}[t]
  \centering
  \includegraphics[width=1\linewidth]{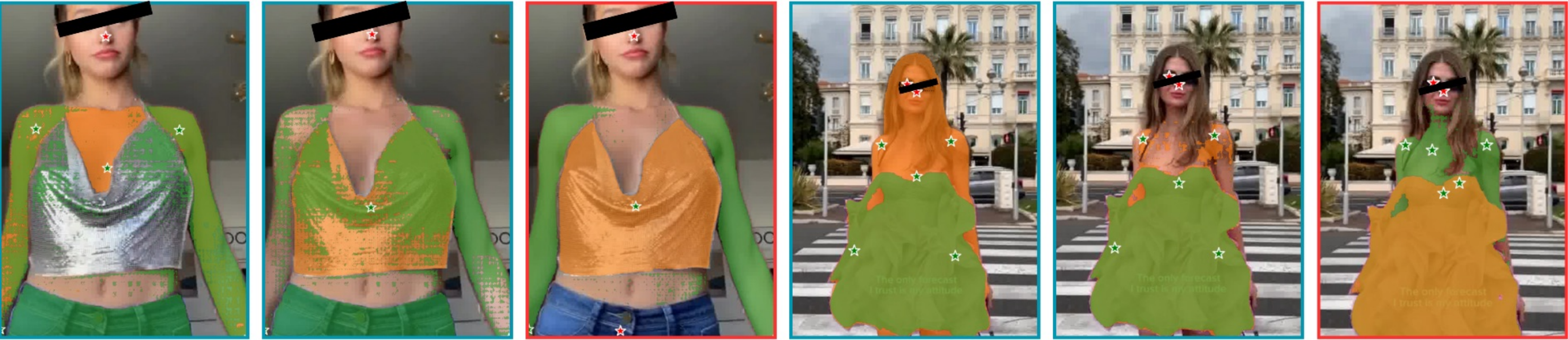}
      \caption{SAM 2 failures due to its sensitivity to prompts (green for positive, red for negative), requiring manual corrections for challenging areas (green rectangles).}

   \label{fig:sam_mask_quality}
\end{figure}

Public datasets for single-image virtual try-ons, such as VITON-HD and DressCode, often suffer from simple backgrounds and limited human poses, despite offering high-quality images. These datasets are also restricted to single-image scenarios. Similarly, the VVT dataset, a standard for video virtual try-on, has notable drawbacks, including uniform movements, white backgrounds, and low resolution ($256\times 192$), making it unsuitable for real-world applications, particularly in the short-video industry where higher resolution is crucial. Since real-world videos are typically recorded on mobile phones, which introduce variations in background, camera position, and lighting, there is a pressing need for a more robust dataset. To address these shortcomings, we introduce \dataname, a high-resolution video virtual try-on dataset that includes complex backgrounds, diverse movements, and balanced gender representation. Each video is paired with its corresponding garment and annotated with detailed human poses and precise binary cloth masks, enhancing its utility for real-world applications.

\textit{First}, the quality of garment masks in our dataset is crucial for enhancing try-on results, as shown in the Supplementary Material. While existing datasets like VITON-HD \cite{choi2021viton}, DressCode \cite{morelli2022dress}, and VVT \cite{dong2019fw} use a standard segmenter~\cite{li2020self}, it struggles with complex videos, resulting in subpar performance. In contrast, \dataname~offers \textbf{manually corrected}, highly accurate garment masks, leading to significantly improved video try-on quality.

\textit{Second}, our \dataname~dataset captures a broad range of human poses and dynamic movements, such as dancing, common in short-form videos. As shown in \cref{fig:dataset_samples}, it includes variations in camera distance and diverse backgrounds, from indoor to outdoor settings with complex lighting. Additionally, it features a variety of clothing types, from casual T-shirts to structured garments like sweaters and challenging attire such as chainmail tops, addressing real-world challenges in video virtual try-on.

\subsection{Video collection and annotation.} 

Our dataset consists of short TikTok clips (10–30 seconds) showcasing various dance routines, as illustrated in Fig.~\ref{fig:dataset_samples}. We expanded the TikTok Dataset \cite{Jafarian_2021_CVPR_TikTok} by adding videos to enhance diversity in backgrounds, skin tones, and clothing styles, resulting in 817 video-garment pairs. Videos with excessive motion blur or low-quality garments were excluded. To ensure accurate garment matching, we manually curated high-quality matches from fashion retail websites. The dataset includes over 270,000 RGB frames extracted at 30 frames per second. Additionally, we computed 2D keypoints and dense pose information using DWPose \cite{yang2023effective} and DensePose \cite{guler2018densepose}. Dataset statistics, summarized in \cref{tab:datastat}, highlight its diversity in gender, skin tone, and camera positions.

Creating high-quality garment masks for each video posed a significant challenge due to the need for precise segmentation in every frame. We used SAM 2 \cite{sam2} to extract masks for both clothing and arms. However, its sensitivity to prompt points and specific frames (see Fig.~\ref{fig:sam_mask_quality}-(a)) necessitated an additional solution. To improve efficiency, we developed an algorithm (detailed in the Supplemental Material) for optimal frame and prompt selection. Complex garments still required manual refinement, as shown in the Supplementary Material. This meticulous process was essential for ensuring the dataset's high quality and reliability.

\section{Experiments}

\myheading{Datasets:} We evaluate our approach on the VVT dataset \cite{dong2019fw} and our new \dataname~dataset. The VVT dataset, a standard benchmark for video virtual try-on, includes 791 paired videos of individuals and clothing images, with 661 for training and 130 for testing, all at $256\times192$ resolution. The videos feature simple movements against plain backgrounds. In contrast, the \dataname~dataset offers a more complex challenge, with varied backgrounds, dynamic movements, and diverse body poses. It comprises 693 training videos and 124 testing videos at $540 \times 720$ resolution, totaling 232,843 frames for training and 39,705 frames for testing.

\myheading{Metrics:} We evaluate our approach using image-based and video-based metrics in both paired and unpaired settings, as outlined in \cite{jiang2022clothformer}. In paired settings, we use SSIM \cite{wang2004image} and LPIPS \cite{zhang2018unreasonable} to assess reconstruction quality. In unpaired settings, we measure visual quality and temporal consistency with Video Fréchet Inception Distance (VFID) \cite{dong2019fw}. Additionally, we measure inference speed in frames per second (FPS) to demonstrate speed improvements.

\myheading{Implementation details:}
The training process is divided into two stages. In the first stage, we focus on inpainting and preserving detailed garment textures using the VITON-HD dataset \cite{choi2021viton}. We fine-tune the Garment UNet, Pose Encoder, and Main UNet decoder, initializing the Main UNet and Garment UNet with pretrained weights from SD 1.5, while keeping the VAE Encoder, Decoder, and CLIP image encoder weights unchanged. In the second stage, we incorporate temporal attention layers into the previously trained model, initializing these new modules with pretrained weights from AnimateDiff \cite{guo2023animatediff}.

\begin{table}[t]
\small
\centering
\setlength{\tabcolsep}{2pt} 
\begin{tabular}{l c c c c c}
  \toprule
  \multirow{2}{*}{\textbf{Method}} & \multicolumn{5}{c}{\textbf{VVT}} \\
  \cmidrule(lr){2-6}
   & \textbf{LPIPS $\downarrow$} & \textbf{SSIM $\uparrow$} & \textbf{VFID$_{\text{I3D}}$ $\downarrow$} & \textbf{VFID$_{\text{RN}}$ $\downarrow$} & \textbf{FPS $\uparrow$} \\
  \midrule
  CP-VTON & 0.535 & 0.459 & 6.361 & 12.100& N/A \\
  FBAFN & 0.157 & 0.870 & 4.516 & 8.690 & N/A \\
  StableVITON & 0.184 &  0.760 & 17.068 &11.254& 0.241 \\
  StableVITON+AA & 0.270 &	0.683	& 12.597 & 3.336 & 1.165 \\
   \midrule  
  FWGAN & 0.283 & 0.675 & 8.019 & 12.150 & N/A \\
  MVTON & 0.068 & 0.853 & 8.367 & 9.702 & N/A \\
  ClothFormer & 0.081 & \textbf{0.921} & 3.967 & 5.048 & N/A \\
   \midrule  
   Tunnel Try-On & \textbf{0.054} & \underline{0.913} & \textbf{3.345} & 4.614 & N/A \\
   ViViD\Cross & 0.119 & 0.829 & 6.788 & \underline{0.853}& \underline{1.409} \\
   WildVidFit & N/A & N/A & 4.202 & N/A & N/A \\
  \midrule
  \Approach~(ours) & \underline{0.066} & 0.887 & \underline{3.589} & \textbf{0.534} & \textbf{2.270} \\  
  \bottomrule
\end{tabular}
\caption{Comparisons on the VVT dataset~\cite{dong2019fw}. \Cross~means our re-evaluation from the provided code. 
}
\label{tab:VVT-results}
\end{table}

\begin{table}[t]
\small
\centering
\setlength{\tabcolsep}{2.5pt} 
\begin{tabular}{l c c c c}
  \toprule
  \multirow{2}{*}{\textbf{Method}} & \multicolumn{4}{c}{\textbf{\dataname}} \\
  \cmidrule(lr){2-5}
  & \textbf{LPIPS $\downarrow$} & \textbf{SSIM $\uparrow$} & \textbf{VFID$_{\text{I3D}}$ $\downarrow$} & \textbf{FPS $\uparrow$} \\
  \midrule
    ViViD\Cross  & 0.129 &  0.824 & 5.638 & 1.409 \\  
  \Approach~w/o ShiftCaching & 0.075 & \textbf{0.891} & \textbf{3.865} & 1.177 \\  
  \Approach~(ours) & \textbf{0.074} & 0.888 & 4.231 & \textbf{2.270} \\  
  
  \bottomrule
\end{tabular}
\caption{Comparisons on the \dataname~dataset. 
} 
\label{tab:combined-results}
\end{table}

\begin{table}[t]
\centering
\small
\setlength{\tabcolsep}{4pt}

\begin{tabular}{c c c c c c}
  \toprule
    \textbf{Variant} & \textbf{LPIPS$\downarrow$} & \textbf{SSIM$\uparrow$} & \textbf{VFID$_{\text{I3D}}$$\downarrow$}  & \textbf{VFID$_{\text{RN}}$$\downarrow$} & \textbf{FPS$\uparrow$} \\ 
  \midrule
  FS & 0.061 & 0.882 & 8.971 & 0.864 & 1.544 \\
  RS & 0.060 & 0.883 & \textbf{8.878} & \textbf{0.853} & 1.544 \\
  FS, P 50\% & 0.060 & 0.883 & 8.932 & 0.887 & \textbf{2.270} \\
  RS, P 50\% & 0.060 & 0.883 & 8.938 & 0.888 & \textbf{2.270} \\
  \bottomrule
\end{tabular}
\caption{Study of our Temporal layers with ShiftCaching. FS: Fixed Shift, RS: Random Shift, P: Partially Computed.}
\label{tab:shift_caching_random_drop}
\end{table}

\begin{table}[t]
\centering
\small
\setlength{\tabcolsep}{4.5pt}

\begin{tabular}{ccccc}
  \toprule
  \textbf{Variant} & \textbf{LPIPS$\downarrow$} & \textbf{SSIM$\uparrow$} & \textbf{VFID$_{\text{I3D}}$$\downarrow$}  & \textbf{VFID$_{\text{RN}}$$\downarrow$} \\ 
  \midrule
  FA & 0.060 & 0.883 & 8.932 & 0.887 \\
  HA & \textbf{0.059} & \textbf{0.886} & \textbf{8.679} & \textbf{0.796} \\
  QA & 0.061 & 0.882 & 8.990 & 0.909 \\
  CA & 0.086 & 0.854 & 13.520 & 5.501 \\
  \bottomrule
\end{tabular}
\caption{Ablation study of our Temporal layers with ShiftCaching. FA: Full Attention, HA: Half Attention, QA: Quarter Attention, CA: Causal Attention.}
\label{tab:attention}

\end{table}

\begin{table}[ht!]
\centering
\small
\setlength{\tabcolsep}{4.5pt}
\begin{tabular}{cccccc}
  \toprule
    \textbf{Variant} & \textbf{LPIPS$\downarrow$} & \textbf{SSIM$\uparrow$} & \textbf{VFID$_{\text{I3D}}$$\downarrow$}  & \textbf{VFID$_{\text{RN}}$$\downarrow$} & \textbf{FPS$\uparrow$} \\
  \midrule
  8 & 0.062 & 0.880 & 9.312 & 1.027 & 0.914 \\
  16 & \textbf{0.061} & \textbf{0.881} & 8.822 &  0.851 & 1.176 \\
  24 &  0.163 & 0.821 & \textbf{8.724} & \textbf{0.800} & \textbf{1.723} \\
  \bottomrule
\end{tabular}
\caption{Ablation study of testing with 8, 16, and 24 frames, with the default training set to 16 frames.}
\label{tab:test_frames_ablation}
\end{table}

\subsection{Comparisons with Prior Approaches}
We compare our approach with other video virtual try-on methods using the VVT and \dataname~datasets. As most methods are closed-source, we rely on reported results and available generated videos for comparison.
For GAN-based methods, we evaluate against FW-GAN \cite{dong2019fw}, MV-TON \cite{zhong2021mv}, and ClothFormer \cite{jiang2022clothformer}. For diffusion-based methods, we compare with Tunnel Try-On \cite{xu2024tunnel}, ViViD \cite{fang2024vivid}, and WildVidFit \cite{he2024wildvidfit}. We re-evaluate ViViD \cite{fang2024vivid} on the VVT dataset, as it is the only method with available inference code and pre-trained weights. Additionally, we compare our model with the image-based virtual try-on method StableVITON \cite{kim2023stableviton}, finetuned on the VVT dataset, in a frame-by-frame manner. We also evaluate a baseline combining StableVITON and AnimateAnyone \cite{hu2023animateanyone}, where StableVITON performs the try-on for individual frames, and AnimateAnyone generates a video based on the source motion.

\begin{figure*}[t]
  \centering
  \includegraphics[width=0.9\linewidth]{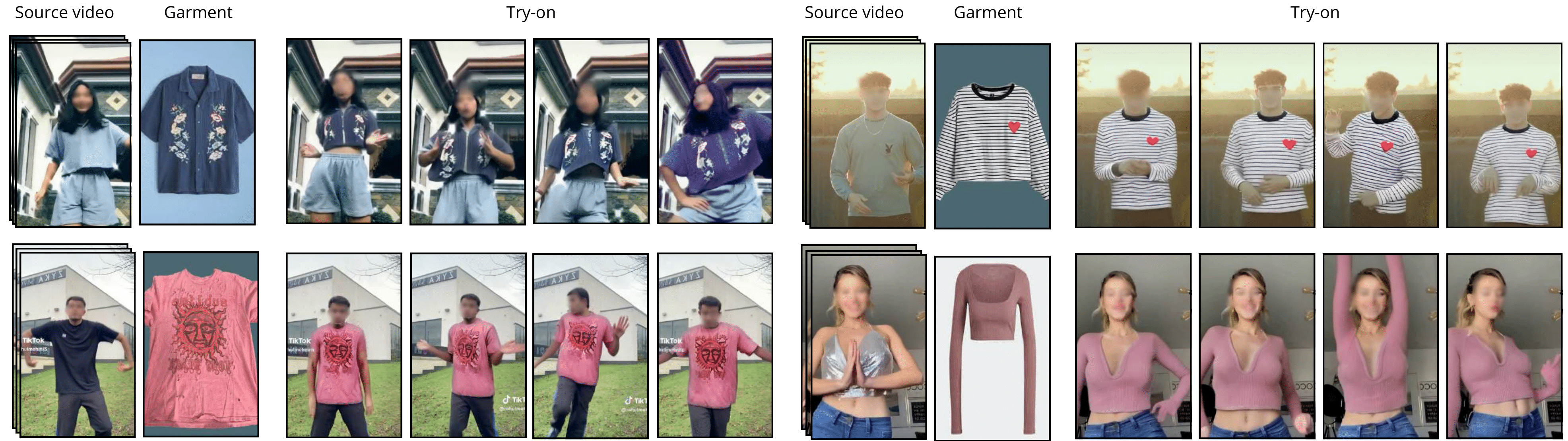}
\caption{Qualitative results of our method on the \dataname~dataset.}
  \label{fig3:qual_tiktol}
\end{figure*}

\begin{figure*}[!h]
  \centering
  \includegraphics[width=0.88\linewidth]{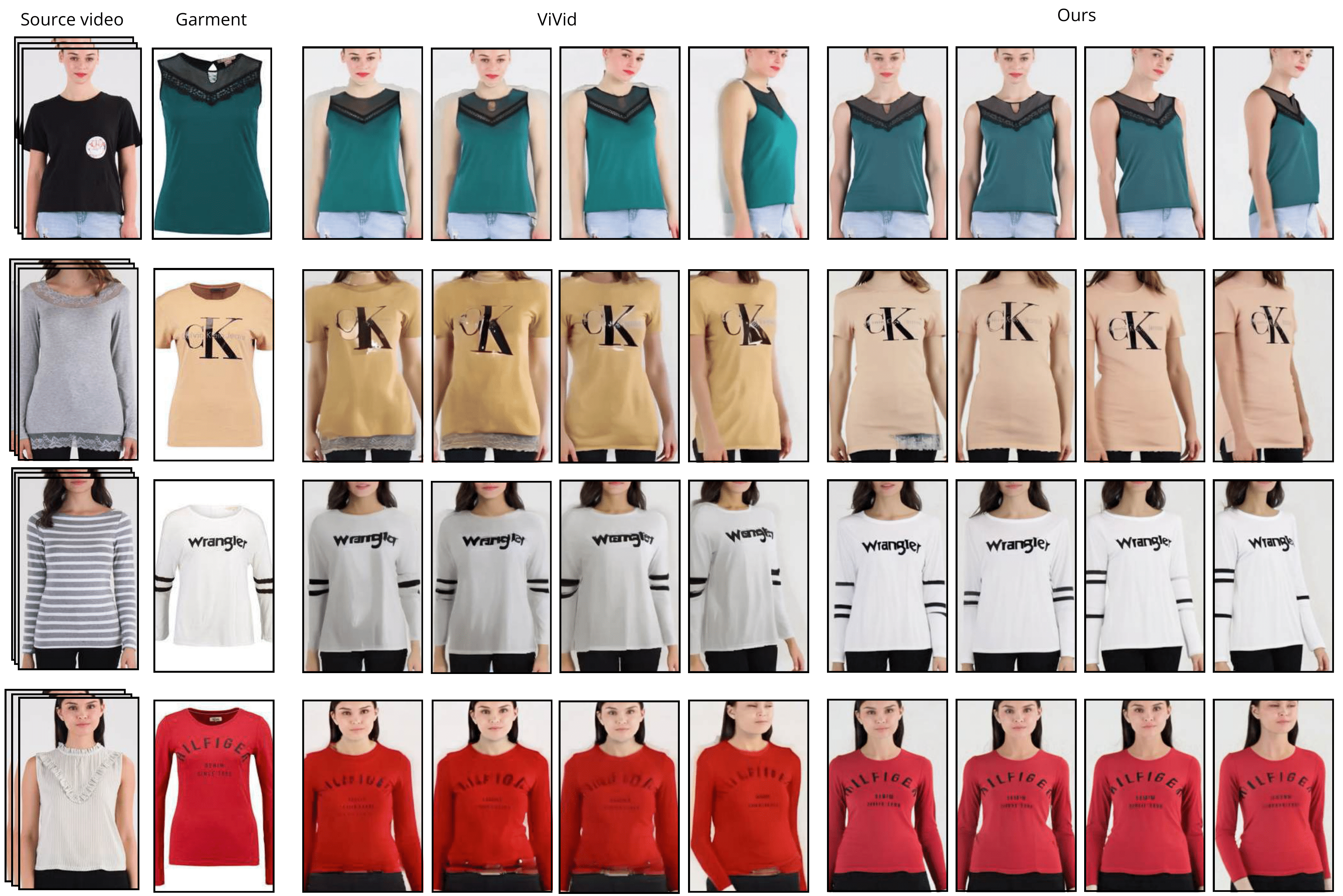}
  \caption{Qualitative comparison with prior method on the VVT dataset.}
  \label{fig4:qual_vvt}
\end{figure*}

\myheading{Quantitative results:} \cref{tab:VVT-results} presents the comparison on the VVT dataset. Our approach excels in the VFID metric, indicating superior visual quality and consistency, and also performs competitively in SSIM and LPIPS scores. While ClothFormer achieves a high SSIM score, its VFID is lower due to the limitations of its GAN-based method. Our ShiftCaching technique enhances performance, increasing the frame rate to 2.27 FPS -- over 1.5 times faster. We also evaluated our method on the \dataname~dataset, as detailed in \cref{tab:combined-results}.
Our analysis shows that while these methods produce accurate individual frames, they often struggle with flickering and inconsistencies due to poor temporal coherence and motion handling across frames.

\myheading{Qualitative results:}
As shown in \cref{fig4:qual_vvt} and \cref{fig3:qual_tiktol}, the textures on the garment vary between frames. Additionally, there are significant jitters between adjacent frames with these methods, which can be observed more intuitively in videos provided in our Supplementary Material.

\subsection{Ablation Study}
We conducted ablation studies on the VVT dataset to investigate various factors affecting the performance of \Approach.

\myheading{Study on the ShiftCaching Technique} is shown in \cref{tab:shift_caching_random_drop}. The results indicate that using random shifts provides the best consistency. When combined with partial computation of 50\% of the frames, this approach accelerates inference by 1.5 times while maintaining comparable quantitative metrics to other methods.

\myheading{Study on Different Types of Masks in Masked Temporal Attention} is shown in \cref{tab:attention}. The results reveal that Half Attention yields the best performance. This suggests that allowing only the bad features (from timestep \( t+2 \)) to access the good features (from timestep \( t+1 \)) and allowing only the good features to interact with each other, produces the optimal results. Detailed explanations of different masking attention are described in our Supplementary Material.

\myheading{Impact of Inference Video Chunk Length} is examined in \cref{tab:test_frames_ablation}. The study reveals that matching the training and inference video chunk lengths -- both set to $N=16$ -- yields the best results.

\section{Conclusion}
In conclusion, we have proposed a novel technique, ShiftCaching, which ensures temporal smoothness across video chunks while effectively reducing redundant computations during video inference. This advancement enhances the efficiency and quality of video virtual try-on, making it more practical for real-world applications. Additionally, we have introduced a new dataset, \dataname, designed specifically for video virtual try-on. This dataset stands out for its diverse range of backgrounds, complex movements, and high-resolution videos, addressing the limitations of existing datasets and providing a valuable resource for future research in this area.

\bibliography{aaai25}
\maketitlesupplementary

\appendix

In this supplementary material, we present additional experimental results and details that could not be included in the main paper due to space constraints. First, we provide a comprehensive overview of the training process for our proposed method. Next, we elaborate on the various masking strategies employed in the attention module. We also include detailed information about the creation of our TikTokDress dataset. Finally, we showcase additional qualitative results of our method on both the VVT and TikTokDress datasets.

\section{Training Detail}
\myheading{Optimization.} We use the Adam~\cite{kingma2014adam} optimizer with $\beta_1=0.9$ and $\beta_2=0.999$, and a fixed learning rate of $1e-5$ for both the image pretraining stage and stage 2. We also adopt the Min-SNR Weighting Strategy~\cite{hang2023efficient} with $\gamma=5.0$. All training is conducted in mixed-precision FP16.

\myheading{Hardware.} The code is implemented in PyTorch using the Diffusers~\cite{von-platen-etal-2022-diffusers} framework, and training is performed on a single A100 GPU with 40GB memory for both stages. The training for stage 1 on VITON-HD~\cite{choi2021viton} takes approximately 3 days, while stage 2 on both the VVT and TikTokDress datasets takes around 30 hours. The FPS benchmarks reported in the paper were also measured using the A100 GPU.

\myheading{Training.} In the first stage, we optimize the Main UNet decoder, Garment UNet, and Pose Encoder on VITON-HD~\cite{choi2021viton}, keeping all other modules fixed. The training resolution for this stage is maintained at $1024\times768$, consistent with the original dataset resolution. In the second stage, we sample a clip of 16 frames with a resolution of $512\times384$ from each video as input. At this stage, we optimize only the motion-related modules.

\section{Different Masking Attention Mechanisms}
\begin{figure}[ht]
    \centering
    \includegraphics[width=0.9\linewidth]{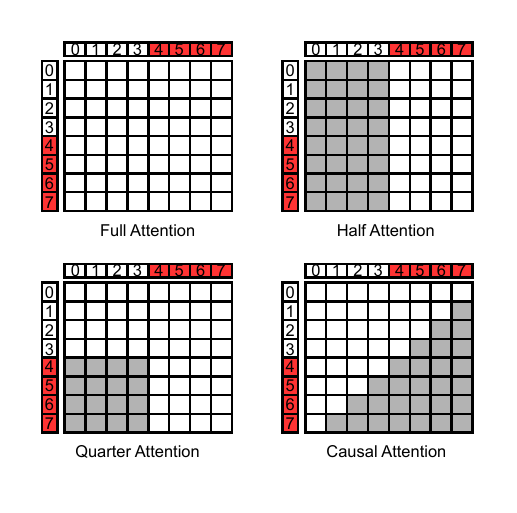}
    \caption{The attention mask used in different masked temporal attention mechanisms in ShiftCaching. White cells in the matrix represent a value of $0$, while gray cells indicate a value of $-\infty$.}
    \label{fig:ablation_attention}
\end{figure}

\cref{fig:ablation_attention} illustrates the different attention masking strategies in our ShiftCaching method. In the Full Attention configuration ($a$), the attention mask is filled with $0.0$, allowing each frame to attend to all others. However, this setup can result in less accurate features (e.g., $t+2$, shown in white) influencing more accurate ones (e.g., $t+1$, shown in red). In the Half Attention setup ($b$), the query frame always attends to the accurate features ($t+1$), enabling the transfer of high-quality features to less accurate ones. The Quarter Attention setup ($c$) allows frames with less accurate features to attend to all others while ensuring that accurate features do not attend to less accurate ones, thereby preserving their quality. Finally, the Causal Attention setup ($d$) ensures that each frame's computation depends only on succeeding frames (less noise frames). Our experiments reveal that Half Attention ($b$) yields the best performance.

\section{\dataname~Dataset}
\cref{fig:dataset_sample} showcases examples from our dataset, highlighting its diversity in backgrounds (ranging from indoor to outdoor settings), camera positions (including top and bottom angles), person distances (ranging from far to close to the camera), skin tones (spanning Black, Asian, and White), and garment types (such as T-shirts, sweaters, crop tops, long-sleeve T-shirts, and dresses). This diversity makes our dataset more representative of real-life scenarios for virtual try-on applications.

\begin{figure}[ht]
    \centering
    \includegraphics[width=1\linewidth]{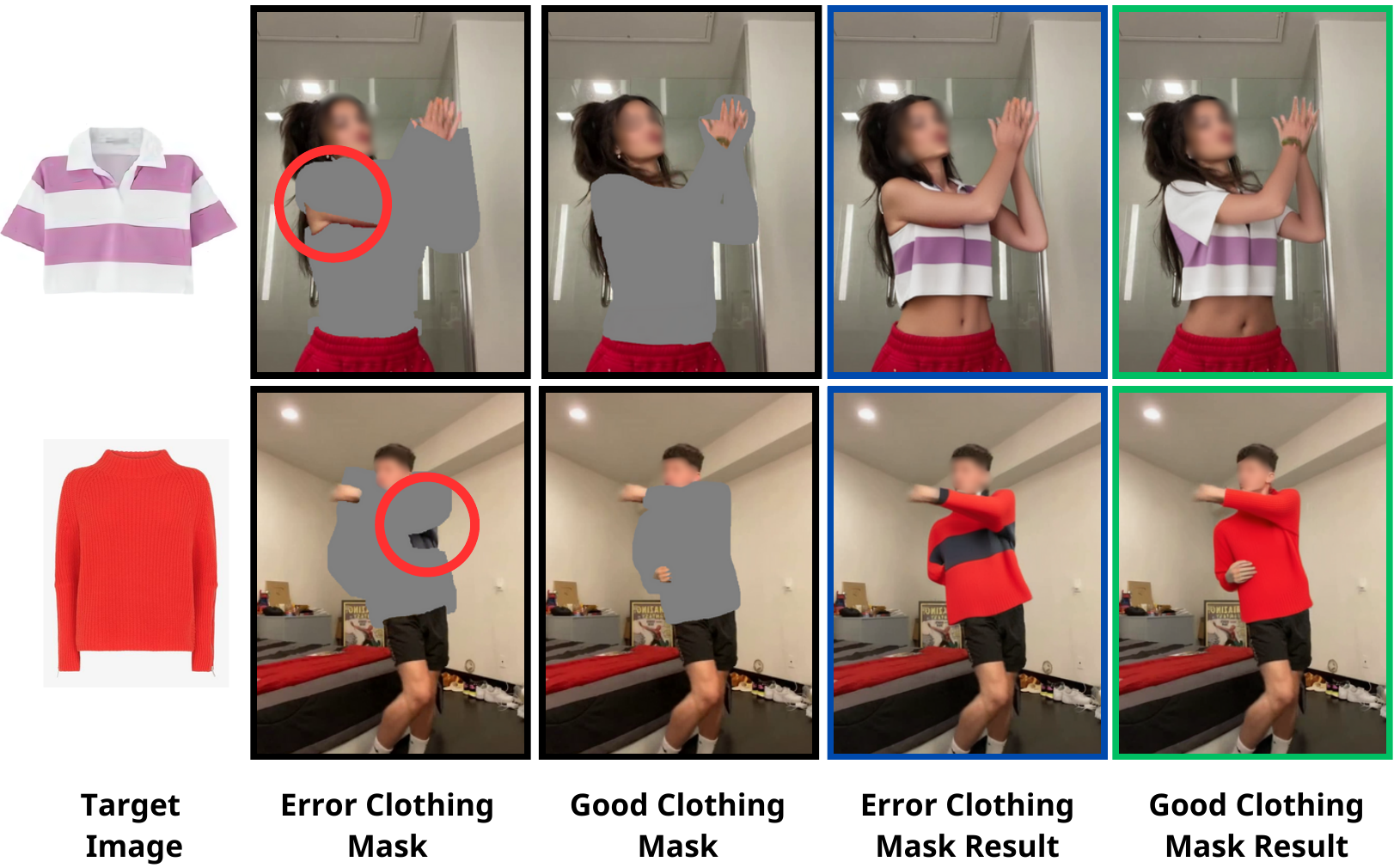}
       \caption{Impact of Mask Quality on Try-On Results. The first row displays the target clothing images. The second row illustrates the effects of an inaccurate mask, resulting in flawed try-on outputs shown in the third row. The fourth row demonstrates how a precise mask significantly enhances the quality and accuracy of the results. Accurate masks are essential for achieving high-quality try-on outcomes.}
    \label{fig:mask_cmp_supp}
\end{figure}
\begin{figure}[!t]
    \centering
    \includegraphics[width=0.5\linewidth]{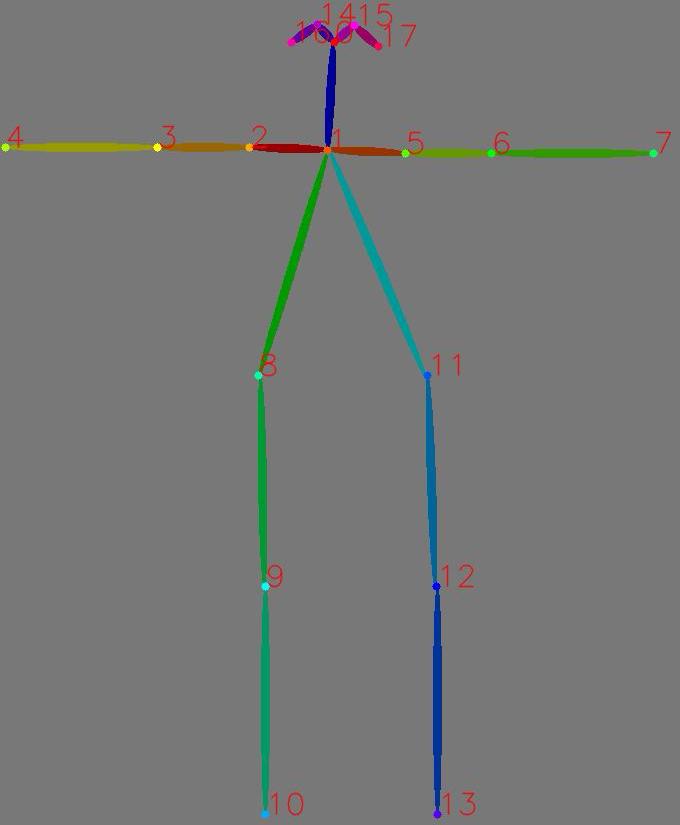}
       \caption{Examples of T-pose}
    \label{fig:t-pose}
\end{figure}

\myheading{Agnostic masks} are essential for achieving accurate try-on results. For instance, in \cref{fig:mask_cmp_supp}, an incorrect agnostic mask (highlighted by the red circle) prompts the model to hallucinate a garment to fit the mask, resulting in a try-on outcome where the garment appears as a tank top (highlighted by the blue rectangle in the first row). Conversely, using a correct garment mask produces a satisfactory try-on result (highlighted by the green rectangle in the first row). Similarly, in the second row, a garment that is not fully masked causes the model to hallucinate additional red stripes in the middle of the red garment (highlighted by the blue rectangle in the second row). Once again, a correct garment mask leads to an accurate try-on result (highlighted by the green rectangle in the second row).

\begin{figure*}[ht]
    \centering
    \includegraphics[width=0.8\linewidth]{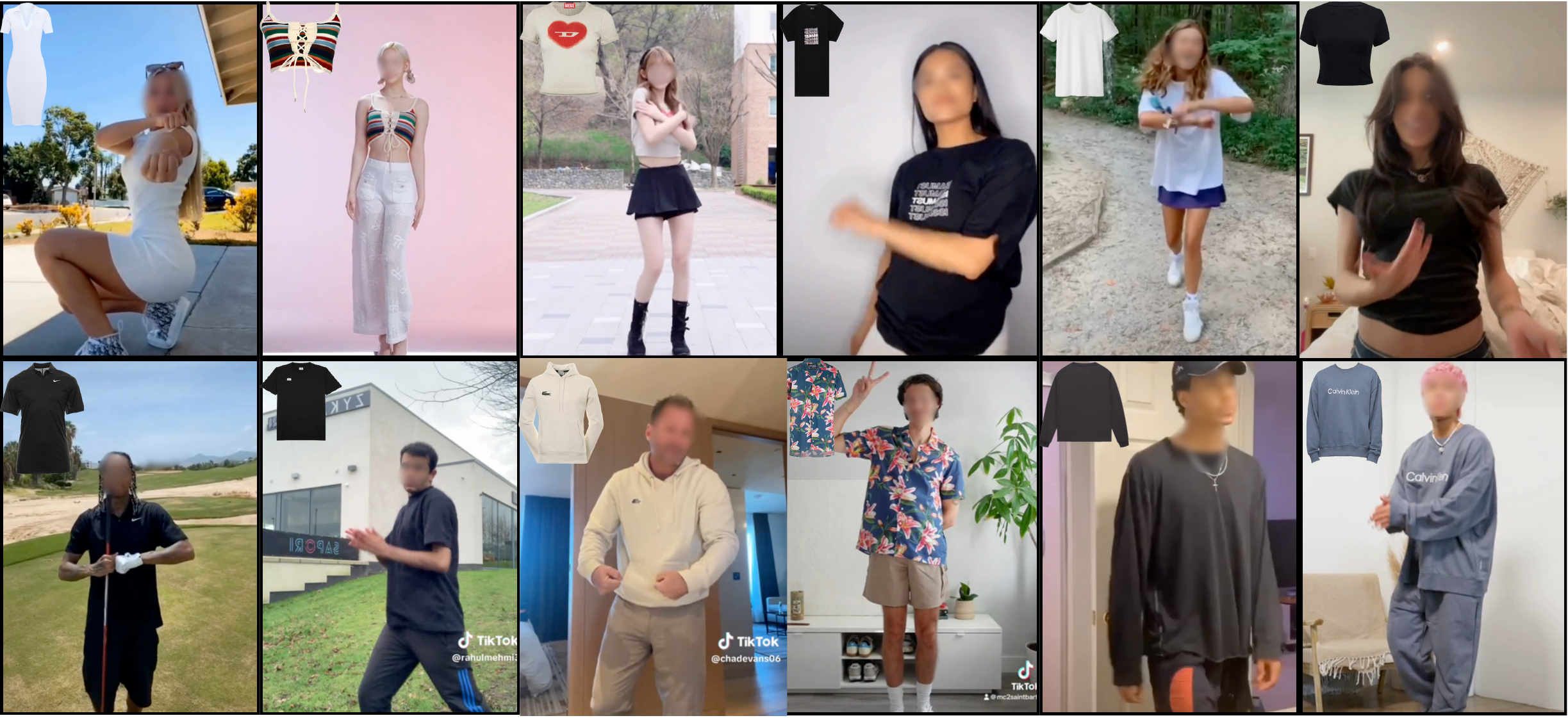}
       \caption{Examples from our dataset, which feature diverse poses and high-quality garments.}
    \label{fig:dataset_sample}
\end{figure*}

To obtain accurate agnostic masks, we use SAM2 \cite{sam2} for both garment and arm segmentation. From a video with multiple frames showing diverse poses, we select the frame that displays the most visible body parts and has the highest clarity, as illustrated in \cref{fig:t-pose}. We apply the algorithm described in \cref{alg:prompting} to facilitate this frame selection.

The agnostic masks are created by combining the garment mask and the arm mask. For garment masks, we select positive points during the prompting process, specifically the body center, right shoulder, and left shoulder. The body center is defined as the midpoint between the neck, right hip, and left hip (refer to points 1, 8, and 11 in \cref{fig:t-pose}). Negative points include noise, the left eye, and the right eye. For arm masks, the positive points selected are the left elbow, right elbow, left wrist, and right wrist, while negative points include the right knee, nose, and left knee. After generating the garment and arm masks, we combine them and apply dilation three to seven times to ensure the mask fully conceals the original garment's shape.

\myheading{Garment Image.} To extract garment images from videos, we use the Google Lens engine to find the best match at the highest possible resolution. If Google Lens does not provide a satisfactory result, we manually search on Google using keywords related to the garment type, color, and brand, selecting the most suitable match. If no appropriate garment image can be found, we exclude the video from our dataset. After obtaining the garment image, we occasionally horizontally flip it to align with the video.

\begin{algorithm}[h]
\caption{Selecting the Optimal Frame for SAM2 Prompting}
\label{alg:prompting}

\begin{algorithmic}[1]
\STATE \textbf{Define:}
\STATE \hspace{1em} \texttt{kps} $\gets$ \texttt{keypoints} 
\STATE \hspace{1em} \texttt{p\_joints} $\gets$ \texttt{pre\_define\_joints}
\STATE \hspace{1em} \texttt{v\_p} $\gets$ \texttt{visible\_part}

\STATE \textbf{Initialize:}
\STATE \hspace{1em} \texttt{pre\_define\_joints} $\gets$ \{\texttt{"left\_shoulder"}: [1, 2, 3], \texttt{"right\_shoulder"}: [1, 5, 6], \dots \}
\STATE \hspace{1em} \texttt{body\_angle} $\gets$ \{\texttt{"left\_shoulder"}: 180, \texttt{"right\_shoulder"}: 180, \dots \}
\STATE \hspace{1em} \texttt{frame\_kps\_scores} $\gets$ []

\STATE \textbf{Function} \texttt{perfect\_pose\_score(keypoints)}:
\STATE \hspace{1em} \texttt{score} $\gets$ 0
\STATE \hspace{1em} \texttt{visible\_part} $\gets$ 0
\STATE \hspace{1em} \textbf{for each} \texttt{key} \textbf{in} \texttt{p\_joints} \textbf{do}:
\STATE \hspace{2em} \texttt{j\_A}, \texttt{j\_B}, \texttt{j\_C} $\gets$ \texttt{p\_joints[key]}
\STATE \hspace{2em} \texttt{angle} $\gets$ \texttt{calculate\_angle(kps[j\_A], kps[j\_B], kps[j\_C])}
\STATE \hspace{2em} \texttt{score} $\gets$ \texttt{score} + $\texttt{abs}(\texttt{angle} - \texttt{body\_angle[key]})$
\STATE \hspace{3em} \texttt{v\_p} $\gets$ \texttt{v\_p} + 1
\STATE \hspace{1em} \textbf{return} \texttt{score}, \texttt{v\_p}

\STATE \textbf{Iterate through frames:}
\STATE \hspace{1em} \textbf{for} \texttt{frame\_idx}, \texttt{kps} \textbf{in} \texttt{frames} \textbf{do}:
\STATE \hspace{2em} \texttt{score}, \texttt{v\_p} 
 $\gets$  \texttt{perfect\_pose\_score(kps)}

\STATE \hspace{2em} \texttt{frame\_kps\_scores.append((score, frame\_idx, v\_p))}

\STATE \textbf{Sort frames by visible parts and score:}
\STATE \hspace{1em} \texttt{frame\_kps\_scores.sort(key=lambda x: (-x[2], x[0]))}

\STATE \textbf{Select the frame with the highest visible parts and perfect body pose score.}
\end{algorithmic}
\end{algorithm}

\section{More Visualization Results}
\subsubsection{Qualitative results on the VVT dataset}

In the video try-on task, we compare our method on the VVT dataset \cite{dong2019fw} with ViViD \cite{fang2024vivid}, as shown in \cref{fig:vvt-res}. In video $b.mp4$ (see Supplemental Videos), our method consistently preserves garment textures throughout the video, outperforming ViViD in maintaining texture stability over time. Similarly, in video $d.mp4$ (see Supplemental Videos), the''GAP'' letters on the garment produced by ViViD exhibit noticeable flickering due to its temporal averaging technique. In contrast, our method, which incorporates the ShiftCaching technique, significantly reduces flickering, ensuring smoother and more stable visual results. \textit{We recommend viewing the supplementary videos for a detailed comparison}.

\begin{figure*}[]
    \centering
    \includegraphics[width=0.8\linewidth]{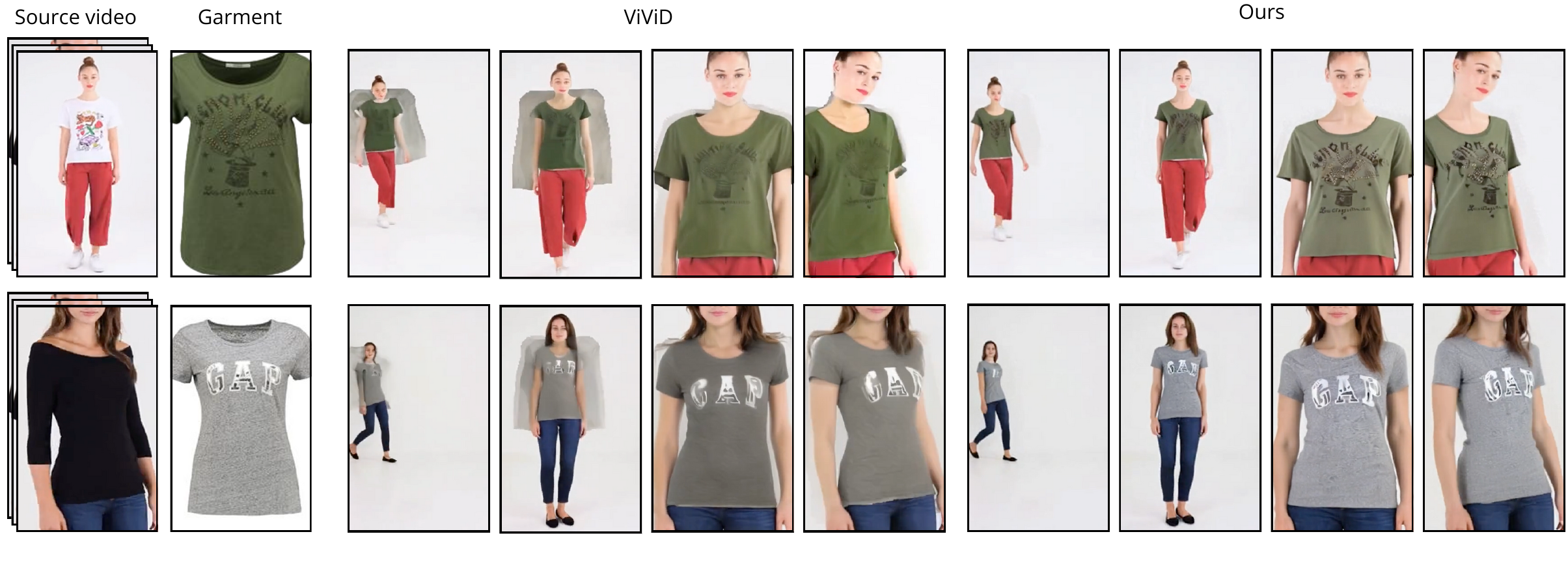}
       \caption{Qualitative comparison between our method and ViViD. Our model achieves robust results over long videos, producing a coherent video sequence.}
    \label{fig:vvt-res}
\end{figure*}

\begin{figure*}[!]
    \centering
    \includegraphics[width=0.7\linewidth]{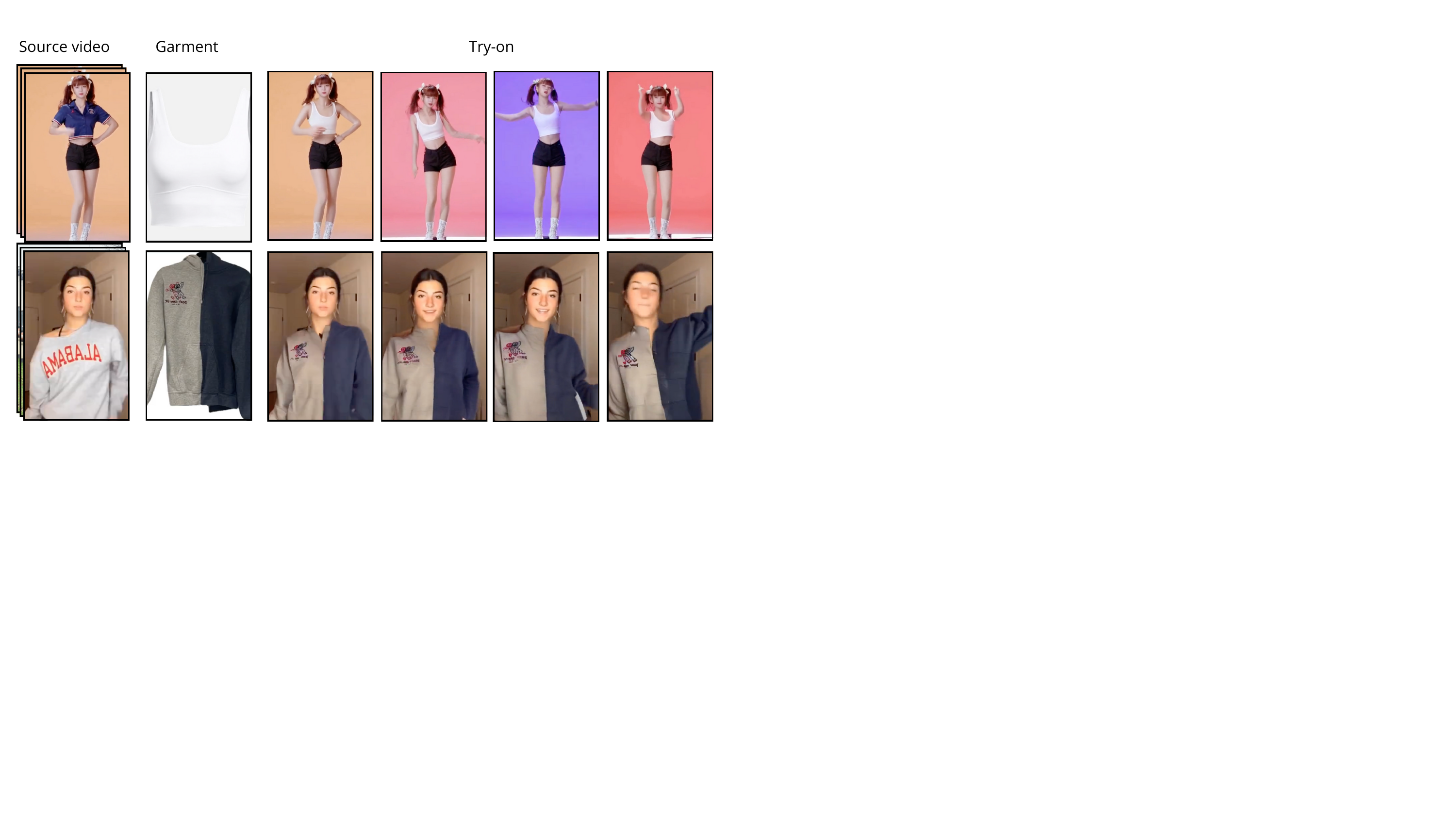}
       \caption{Additional examples of our video try-on results on the TikTokDress dataset. The first row illustrates how effectively our method performs across various backgrounds, while the second row highlights its ability to preserve texture consistency over time.}
    \label{fig:tiktok-res}
\end{figure*}

\subsubsection{Qualitative results on the \dataname~dataset.}
\cref{fig:tiktok-res} showcases additional results of our video virtual try-on method on the \dataname~dataset. By leveraging SAM 2 \cite{sam2} and manually refining masks in our dataset, combined with the SAM 2 frame selection algorithm (\cref{alg:prompting}), our try-on results demonstrate superior performance compared to those trained on the VVT dataset. Our method ensures consistent garment texture quality and, when paired with ShiftCaching, effectively generates continuous and smooth long videos. \textit{More details can be found in the supplementary videos).}

\end{document}